\documentclass{article}
\usepackage{iclr2027_conference,times}

\usepackage{amsmath,amsfonts,bm}

\def\eqref#1{equation~\ref{#1}}

\def\1{\bm{1}}

\DeclareMathAlphabet{\mathsfit}{\encodingdefault}{\sfdefault}{m}{sl}
\SetMathAlphabet{\mathsfit}{bold}{\encodingdefault}{\sfdefault}{bx}{n}

\usepackage{hyperref}
\usepackage{url}

\usepackage{booktabs}
\usepackage{graphicx}
\usepackage{amsmath}

\newcommand{\method}{ProactGen}

\usepackage{fontspec}
\usepackage{framed}
\newenvironment{promptbox}{\begin{framed}\small\raggedright\sloppy\setlength{\parskip}{0.4em}}{\end{framed}}

\title{ProactGen: Learning Proactive\\Clarification for Video Generation}

\author{Zhixiao Wang\textsuperscript{1}, Yang Chen\textsuperscript{1}, Yifan Xu\textsuperscript{1}, Chenbin Li\textsuperscript{2}, Yue Chang\textsuperscript{3}, \\
\bf Kai Long\textsuperscript{4}, Fengming Dong\textsuperscript{1}, Limin Wang\textsuperscript{1}\setcounter{footnote}{1}\thanks{Corresponding author.} \\
\textsuperscript{1}Nanjing University \\
\textsuperscript{2}University of Science and Technology of China \\
\textsuperscript{3}The Hong Kong University of Science and Technology (Guangzhou) \\
\textsuperscript{4}Shanghai Jiao Tong University
}

\iclrfinalcopy
\begin{document}

\maketitle
\lhead{}
\renewcommand{\headrulewidth}{0pt}
\begin{abstract}
Recent advances in video generation let users create high-quality videos from natural language descriptions, yet obtaining results that match user intent remains time-consuming: individual generations are computationally expensive, and underspecified prompts force repeated cycles of generation, inspection, and revision. Existing video prompt optimization methods improve generation through automatic rewriting, but cannot supply the ideas and preferences that were not in the initial description. We propose \method{}, a video generation framework that learns proactive clarification, eliciting user intent through questions grounded in generated video drafts. Specifically, we train a vision-language model as a questioning policy with reinforcement learning: observing the initial prompt and a video draft, it learns to ask the questions most valuable to downstream generation, rewarded by the alignment between the final generated video and a target reference. We further find that effective questioning does not require drafts of final-output quality, and exploit this property by generating drafts with fewer denoising steps, keeping the added cost of clarification low. Experiments show that \method{} improves target alignment over direct generation by at least $1.8\times$ the gain of the strongest baselines. Moreover, a single round of \method{} already exceeds the alignment that the strongest baseline reaches after four refinement rounds, in 29\% of its time.
\end{abstract}

\section{Introduction}
\label{sec:intro}

Rapid progress in video generation has enabled ordinary users to create high-quality videos from a single natural language description~\citep{wan2025wan,kong2024hunyuanvideo,yang2024cogvideox}. Turning that description into a video that matches what the user had in mind, however, remains time-consuming, for two compounding reasons. First, video generation is expensive: every full generation runs tens of denoising steps over a spatiotemporal latent space and costs substantial computation and waiting time. Second, users rarely manage to specify everything they want in the initial prompt. Details such as the subject's appearance, the manner of an action, the lighting, or the camera motion are often recognized as missing or misinterpreted only after the user sees the result~\citep{hutchinson2022underspecification,hahn2025proactive}. Together, these difficulties force users through repeated cycles of generation, inspection, and revision, each paid for with a costly full generation.

\begin{figure}[t]
  \centering
  \includegraphics[width=\linewidth,trim=83 97 65 75,clip]{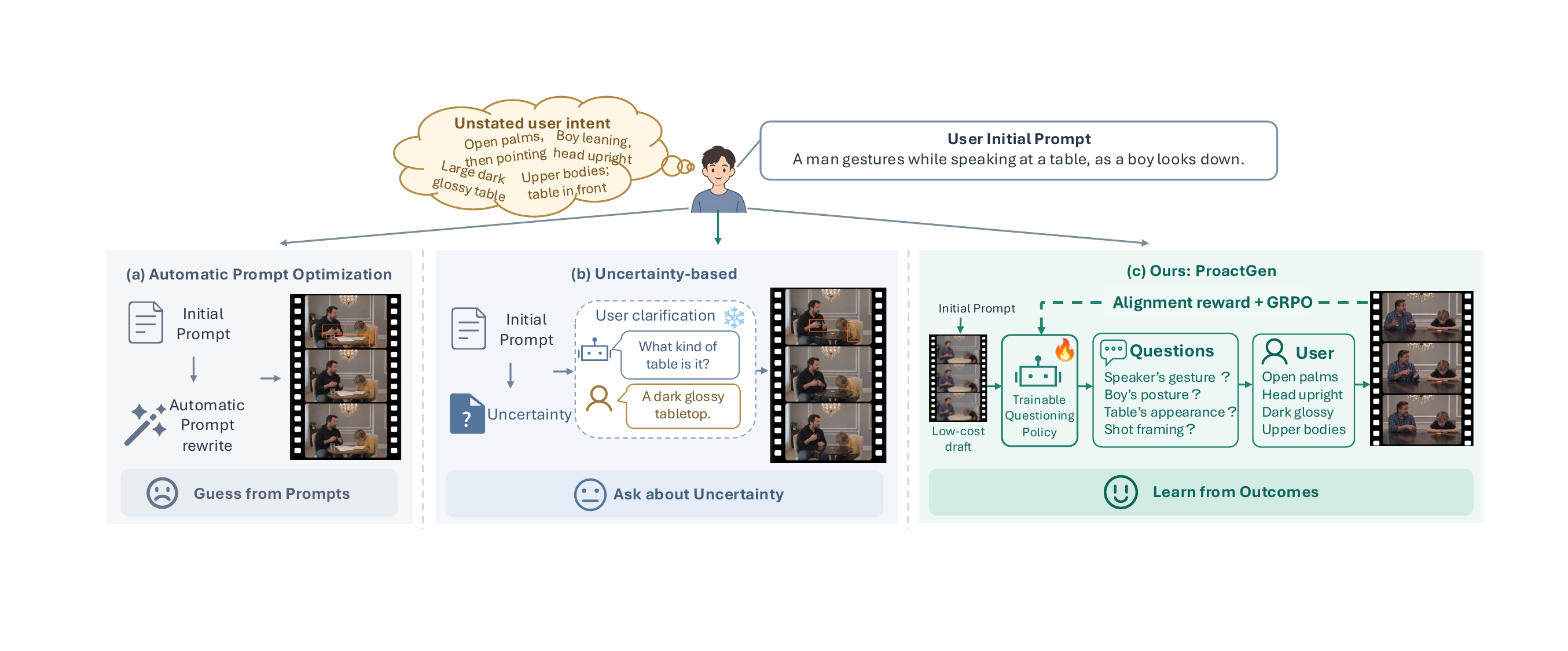}
  \caption{A user's initial prompt often omits preferences the user already has in mind (here, the hand gesture, the boy's posture, the table's material, and the framing). (a) Automatic prompt optimization rewrites the prompt from the prompt alone, so missing preferences can only be guessed. (b) Uncertainty-based clarification asks the user, but derives its questions from the prompt text without knowing what the generator will actually produce. (c) \method{} first observes what the generator produces for the prompt through a low-cost draft, asks the user targeted questions grounded in it, and learns what to ask from the alignment of the final video.}
  \label{fig:teaser}
\end{figure}

Existing work addresses this problem mainly by rewriting the prompt. Video prompt optimization methods~\citep{gao2025rapo,cheng2025vpo,ji2024promptavideo,yang2026scmapr} use language models to expand or rewrite the initial prompt into a form the generator handles well, and some further refine the rewrite with feedback from generated videos~\citep{gao2025rapopp,long2025vista}. These methods improve how the available information is expressed, but cannot supply the ideas and preferences that were never in the initial description: the generator has no way of knowing what color the user pictured for the subject or whether the camera should push in or orbit, and a rewrite can only substitute a guess. A second line of work lets the system interact with the user: in image generation, users express preferences through editing instructions or by choosing among candidates~\citep{wang2024diffchat,huang2024dialoggen,nabati2024pasta,li2026appo}, or the system proactively asks clarifying questions~\citep{hahn2025proactive} derived from a belief state parsed from the prompt; in language modeling, clarification questions have been trained with the quality of the eventual answer as the signal~\citep{andukuri2024stargate,zhang2024futureturns}. These results show that interaction can recover intent that rewriting cannot, and that which question to ask can be learned from what happens afterwards. In these settings, however, questions are selected from the text of the request or a belief state parsed from it, and learning, where present, is supervised by a text response or by the user's choice among candidates. How to choose questions from what a video generator actually produces for a given prompt, and to learn that choice from the final video, remains open; to our knowledge, proactive clarification has not been studied in video generation.

We propose \method{}, a video generation framework that learns proactive clarification: its questions are grounded in a generated draft, and what to ask is learned from the final generation outcome. Given the initial prompt, the generator first produces a video draft; a questioning policy observes both the prompt and the draft and asks a small number of scoped questions; the user's answers are merged into a refined prompt, from which the generator produces the final video. The policy is trained with GRPO~\citep{shao2024deepseekmath}, rewarding the alignment between the final video and a target reference, so it learns to read the defaults and uncertainties the draft exposes and to ask for the information that matters most to downstream generation.

Questioning needs a draft, and a draft is itself a video generation; if clarification had to wait for a full-quality generation, its benefit to the user's total time would be largely offset. We observe that effective questioning does not require a draft of final-output quality: questions depend on how the draft renders the subject, layout, action, and camera, and diffusion models settle these aspects early in denoising, with later steps mainly refining texture and detail~\citep{balaji2022ediffi}. \method{} therefore generates drafts with a greatly reduced number of denoising steps and spends the full budget only on the final video. What the draft gives up is the late-stage detail that questioning does not depend on, so clarification adds little beyond a low-cost draft to direct generation while the cues that questioning relies on are largely preserved.

We evaluate \method{} with a simulated user whose intent is given by a hidden reference video. On 500 samples from InstanceCap~\citep{fan2024instancecap}, \method{} achieves higher target alignment than direct generation, representative video prompt optimization methods, and a proactive clarification method adapted to video on all three metrics, and its gain over direct generation is at least $1.8\times$ that of the strongest baselines. Ablations show that reinforcement learning brings a clear gain and that the benefit of observing the draft emerges only after training. Measured in time, one round of \method{}, about 16\,s, already exceeds the alignment reached by four rounds of automatic rewriting or 15 turns of proactive clarification, and further rounds add further gains.

Our contributions are as follows:
\begin{itemize}
  \item
  To our knowledge, \method{} is the first to bring proactive clarification to video generation; we build a reproducible simulated-user protocol and compare with prompt optimization and proactive clarification baselines.
  \item
  We propose a reinforcement learning framework in which the questioning policy learns, from the final generation outcome, what to ask about a generated draft, targeting the defaults the generator fills in on the sample at hand.
  \item
  We find that effective questioning does not need final-quality drafts and use few-step drafts for clarification, which at equal time buys more alignment than rewriting.
\end{itemize}

\section{Related Work}
\label{sec:related}

\paragraph{Prompt optimization.}
One line of work improves generation by automatically rewriting the user's prompt. In image generation, Promptist~\citep{hao2023promptist} trains a rewriter with reinforcement learning using aesthetic and relevance scores of the generated images as reward, and RePrompt~\citep{wu2025reprompt} adds chain-of-thought reasoning supervised by multiple reward models. In video generation, RAPO~\citep{gao2025rapo} retrieves modifiers from the training corpus and trains a branch selector, VPO~\citep{cheng2025vpo} and Prompt-A-Video~\citep{ji2024promptavideo} train rewriters from video-level preferences or reward-model signals, and SCMAPR~\citep{yang2026scmapr} runs a multi-agent pipeline of scenario routing, rewriting, and text-level semantic verification. Recent methods further refine the prompt at inference with feedback from generated videos: RAPO++~\citep{gao2025rapopp} inspects the output with a vision-language model and corrects the prompt, and VISTA~\citep{long2025vista} self-improves through a loop of multi-agent critique. Whether or not they look at generated output, the input of these methods is limited to what the user already wrote, and for preferences absent from the prompt they can only fall back on the model's prior. \method{} is complementary: its learned action acquires the missing information from the user.

\paragraph{Interactive clarification.}
A second line of work lets the system interact with the user to obtain intent. In language modeling, clarification questions have been trained with the quality of the downstream answer as the signal~\citep{andukuri2024stargate,dou2025togate,zhang2024futureturns}, or ambiguity has been detected from disagreement among generated code solutions~\citep{mu2023clarifygpt}, with text as the downstream product. In image generation, Proactive T2I~\citep{hahn2025proactive} has a language model parse entities, attributes, and relations from the prompt into a belief graph, selects uncertain and important elements to ask the user about, and proposes an automatic evaluation protocol with a simulated user driven by a reference image; its questioning policies are heuristic or prompted. PASTA~\citep{nabati2024pasta} learns with offline reinforcement learning which set of candidate prompts to present to the user, APPO~\citep{li2026appo} lets the user choose among rendered candidates and optimizes the prompt accordingly, and DiffChat~\citep{wang2024diffchat} and DialogGen~\citep{huang2024dialoggen} revise prompts from user editing instructions or hold multi-turn dialogue. In the video domain, IC-Seg~\citep{yang2026icseg} uses hierarchical GRPO to learn when and how to clarify in referring segmentation. In \method{}, questions are grounded in a video draft the generator produced for the sample at hand, and question selection is learned with the alignment of the final generated video to the target as the reward.

\section{Method}
\label{sec:method}

\method{} lets the system proactively obtain, within a single generation pipeline, the information missing from the user's initial prompt, so that the final video comes closer to what the user intends. We state the problem setup (\S\ref{sec:method:setup}), then describe the interaction pipeline (\S\ref{sec:method:pipeline}), reinforcement learning of the questioning policy (\S\ref{sec:method:rl}), and low-cost drafts (\S\ref{sec:method:draft}).

\subsection{Problem Setup}
\label{sec:method:setup}

The user's intent corresponds to a target video $v^\star$, of which the user provides only an initial prompt $p_0$. The prompt is typically incomplete: it names the subject and the gist but leaves out details such as appearance, scene, the manner of an action, or camera motion, which a frozen text-to-video generator $G$ can only fill in with default guesses when generating $G(p_0)$ directly. We want the system to ask the user a few questions before the final generation, complete the prompt with the answers, and thereby bring the final video $\hat v$ closer to $v^\star$. To train and evaluate this ability reproducibly, we follow the simulated-user protocol of proactive T2I~\citep{hahn2025proactive}: a reference video plays the role of $v^\star$, and a simulated user with access to it answers the questions. The reference is visible only to the simulated user and the reward computation; neither the questioning policy nor the generator can access it, and both see only $p_0$ and the draft the system generates. The evaluation measure is $\mathrm{Align}(\hat v, v^\star)$, the visual alignment between final video and reference (\S\ref{sec:method:rl}).

\subsection{Draft-Grounded Clarification Pipeline}
\label{sec:method:pipeline}

\begin{figure}[t]
  \centering
  \includegraphics[width=\linewidth,trim=155 108 113 49,clip]{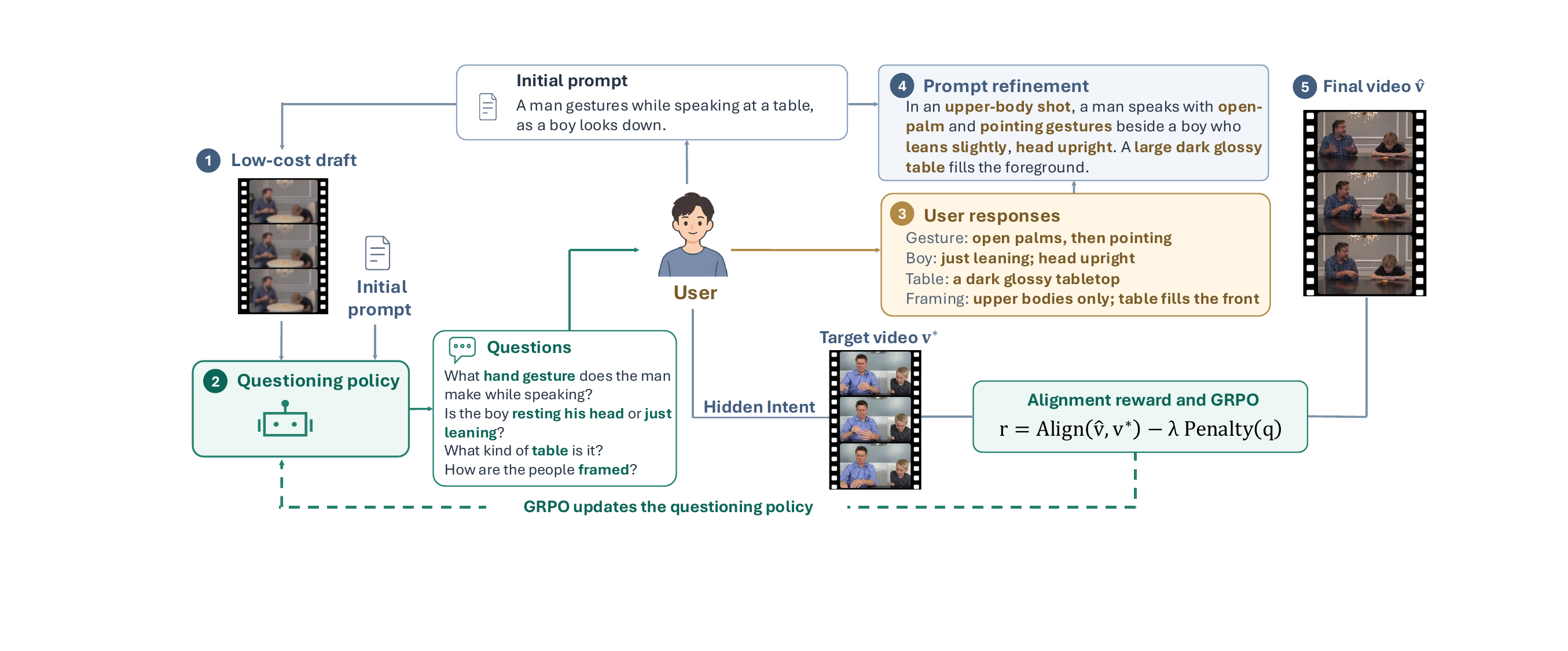}
  \caption{Overview of \method{}. The questioning policy asks scoped questions conditioned on the initial prompt and a low-cost video draft; a simulated user answers from the reference video; the answers are merged into a refined prompt from which the final video is generated at full budget. During training, the alignment between the final video and the reference is the reward for updating the questioning policy; all other components are frozen.}
  \label{fig:overview}
\end{figure}

Figure~\ref{fig:overview} shows the interaction pipeline of \method{}. Given $p_0$, it proceeds in four steps.

\textbf{Draft generation.} The generator produces a video draft $d = G_{\text{draft}}(p_0)$. The draft shows the questioning policy what the generator does with this prompt: which details it fills in by default and where its output departs from the description. It uses a greatly reduced denoising budget (\S\ref{sec:method:draft}).

\textbf{Questioning.} The questioning policy $\pi_\theta$ is a vision-language model that takes $p_0$ and several frames uniformly sampled from the draft and outputs a small set of questions $q \sim \pi_\theta(\cdot \mid p_0, d)$. Each question targets a single piece of information, such as the appearance of an object, the manner of an action, or the camera motion; the policy may neither bundle unrelated items into one question nor request a full description of the video. These constraints keep questions answerable by a real user and prevent reward hacking by asking for the entire description at once.

\textbf{Answering.} The simulated user answers each question briefly from the reference video, under constraints: it addresses only what is asked and adds no facts beyond it, and leaves unanswered any question that falls outside the agreed scope (marked invalid) or cannot be verified from the reference (marked unknown). Answer length is bounded, which together with the constraints on questioning keeps a single exchange from yielding a full description.

\textbf{Refinement and final generation.} A frozen language model merges $p_0$ with all valid question-answer pairs into a refined prompt $p_1$, keeping only content supported by $p_0$ or the answers; the generator then produces the final video $\hat v = G(p_1)$ at full budget.

Only $\pi_\theta$ is learnable in this pipeline; the generator, the simulated user, and the refinement model are frozen. This attributes the learning signal to what was asked and lets \method{} pair with an off-the-shelf video generator without modifying it.

The pipeline can be iterated. When the user wants further correction, the policy observes a new draft generated from the refined prompt $p_1$, or the final video of the previous round, asks again with $p_1$, and the answers are merged into $p_1$ to give $p_2$. Every round follows the same four steps, so the policy needs no separate training for multiple rounds. The method itself does not prescribe whether later rounds observe a draft or a full video; in \S\ref{sec:exp:draft} we evaluate multi-round interaction where every round observes a low-cost draft and only the last runs full generation.

\subsection{Learning to Ask from Generation Outcomes}
\label{sec:method:rl}

We want the questioning policy to select the questions most valuable to downstream generation, a value that can only be observed after the answers are merged and the final video generated. We therefore treat the pipeline as the environment, questioning as the action, and the final video's alignment as the reward, and train $\pi_\theta$ with reinforcement learning.

\textbf{Reward.} We measure alignment with DreamSim~\citep{fu2023dreamsim}, a perceptual image similarity learned from human judgments that agrees better with human notions of whether two images look alike than pixel or low-level feature distances. DreamSim returns a distance $d_{\mathrm{DS}}\in[0,2]$, and we use $1-d_{\mathrm{DS}}$ as the similarity. To apply it to video, we compute this similarity between corresponding frames at $K$ temporally aligned positions of the final video and the reference and average:
\begin{equation}
  \mathrm{Align}(\hat v, v^\star) = \frac{1}{K}\sum_{k=1}^{K}\big[1 - d_{\mathrm{DS}}(\hat v_{t_k}, v^\star_{t_k})\big],
  \label{eq:align}
\end{equation}
Evaluation uses a more densely sampled, temporally aligned version of the same similarity (\S\ref{sec:exp:setup}). Questions the simulated user marks as invalid incur a penalty, which discourages asking outside the agreed scope:
\begin{equation}
  r(q) = \mathrm{Align}\big(G(p_1), v^\star\big) - \lambda\, \mathrm{Penalty}(q),
  \label{eq:reward}
\end{equation}
where $\mathrm{Penalty}(q)$ counts the invalid questions, $\lambda$ is the penalty coefficient, and the total penalty per sample is capped.

\textbf{Optimization.} We use GRPO~\citep{shao2024deepseekmath}. For each training sample $(p_0, d, v^\star)$, we draw $K$ question sets $\{q^{(1)}, \dots, q^{(K)}\}$ from the current policy; each independently goes through answering, refinement, and final generation to obtain its reward $r^{(i)}$. Rewards are standardized within the group into advantages $A^{(i)} = (r^{(i)} - \bar r) / \sigma_r$, and the policy is updated with a clipped importance-ratio objective under a KL penalty toward the base model. Candidates in a group share the same initial noise for final generation, which reduces the effect of noise randomness on reward differences, and the within-group advantage removes differences in the absolute level of alignment across samples.

\textbf{Training data.} Training samples are built from real videos with detailed captions: the video serves as the reference $v^\star$, and $p_0$ is obtained by masking several information slots in the caption and rewriting the rest. Such a $p_0$ keeps the subject and the gist but lacks several details, mirroring the incompleteness of real user prompts, and guarantees that every training sample contains missing information that can be asked for. At evaluation no masking is applied, and all additional information from the simulated user comes from the reference video and its caption.

\subsection{Low-Cost Drafts for Clarification}
\label{sec:method:draft}

The pipeline above needs a draft before asking, and a draft is itself a video generation; if it used the same budget as the final generation, clarification would double the total generation cost, working against the goal of getting the user a satisfactory video sooner. We observe that what questioning needs from a draft differs from what the user needs from the final video: diffusion models largely form the subject, layout, action, and camera early in denoising and mainly refine texture and detail later~\citep{balaji2022ediffi}, and it is the former that questioning relies on. A draft from far fewer steps is therefore visibly coarse, yet already exposes how the generator interprets $p_0$.

\method{} therefore generates drafts with few denoising steps and keeps the full budget for the final generation. The few-step draft uses a timestep schedule different from that of full generation, allocating more of its steps to the high-noise phase where content and layout are decided. The policy is trained on low-cost drafts from the start, so it learns to extract useful information from a coarse draft. Since the final generation budget is unchanged, the final generation itself is unaffected by the number of draft steps. We verify the effect of low-cost drafts on alignment and generation cost in \S\ref{sec:exp:draft}.

\section{Experiments}
\label{sec:exp}

Our experiments answer four questions. (1) Does \method{} bring the final video closer to the user's target than direct generation, video prompt optimization, and a proactive clarification method adapted to video (\S\ref{sec:exp:main})? (2) Does reinforcement learning produce better question selection, and what role does observing the draft play (\S\ref{sec:exp:ablation})? (3) At equal time, does \method{} buy more alignment than other methods, and what do low-cost drafts contribute (\S\ref{sec:exp:draft})? (4) What does the trained policy actually ask (\S\ref{sec:exp:analysis})?

\begin{table}[t]
  \centering
  \caption{Target alignment. Left: 500 InstanceCap samples, measured as similarity between the final video and the reference video. Right: MiraBench-500, measured as ViCLIP alignment between the final video and the structured captions of the reference (Obj., Bkg., Style, Cam.) and its dense caption (Overall). R gives refinement rounds for SSPO, question turns for Proactive T2I, and interaction rounds for \method{}. $^*$The official MiraBench evaluation set is not released; we sample 500 MiraData videos disjoint from our training set and use its metrics on them.}
  \label{tab:main}
  \footnotesize
  \setlength{\tabcolsep}{2.5pt}
  \begin{tabular*}{\linewidth}{@{\extracolsep{\fill}} l c ccc ccccc @{}}
    \toprule
    & & \multicolumn{3}{c}{InstanceCap-500} & \multicolumn{5}{c}{MiraBench-500$^*$} \\
    \cmidrule(lr){3-5} \cmidrule(lr){6-10}
    Method & R & DS-DTW $\uparrow$ & 3D-VAE $\downarrow$ & DnS $\uparrow$ & Obj. $\uparrow$ & Bkg. $\uparrow$ & Style $\uparrow$ & Cam. $\uparrow$ & Overall $\uparrow$ \\
    \midrule
    Direct & -- & 0.658 & 144.03 & $-0.689$ & 0.175 & 0.135 & 0.144 & 0.141 & 0.198 \\
    SCMAPR & -- & 0.657 & 142.94 & $-0.685$ & 0.178 & 0.141 & 0.149 & 0.143 & 0.206 \\
    SSPO & 1 & 0.665 & 143.00 & $-0.676$ & 0.182 & 0.133 & 0.148 & 0.144 & 0.205 \\
    SSPO & 4 & 0.669 & 142.53 & $-0.672$ & 0.182 & 0.133 & 0.146 & 0.143 & 0.204 \\
    Proactive T2I & 5 & 0.662 & 143.55 & $-0.685$ & 0.195 & 0.152 & 0.154 & 0.146 & 0.221 \\
    Proactive T2I & 15 & 0.668 & 143.37 & $-0.678$ & \textbf{0.196} & 0.154 & 0.155 & 0.147 & 0.222 \\
    \midrule
    \method{} & 1 & \textbf{0.678} & \textbf{142.34} & $\mathbf{-0.659}$ & 0.192 & \textbf{0.162} & \textbf{0.158} & \textbf{0.149} & \textbf{0.225} \\
    \bottomrule
  \end{tabular*}
\end{table}

\subsection{Experimental Setup}
\label{sec:exp:setup}

\textbf{Task and data.} We evaluate text-to-video generation under the simulated-user protocol of \S\ref{sec:method:setup}, with two evaluation sets. The first is 500 videos from InstanceCap~\citep{fan2024instancecap}, with $p_0$ the original caption without masking. The second is built on MiraData~\citep{ju2024miradata}: since the official MiraBench evaluation set is not released, we sample 500 MiraData videos disjoint from our training set, use the short captions that come with the dataset as $p_0$ without rewriting, and evaluate with the text--video alignment metrics of MiraBench (below); we denote this set MiraBench-500. The training set comprises about 15K MiraData videos with detailed captions, disjoint from both evaluation sets; the training $p_0$ is obtained by masking 2 to 5 information slots (appearance, scene, lighting, action, camera, and so on) in the caption and rewriting the rest with a language model.

\textbf{Models and implementation.} The generator is Wan2.2-TI2V-5B~\citep{wan2025wan}, producing 49-frame videos at 832$\times$480; final generation uses 20 denoising steps, and drafts use 8 during both training and testing. The questioning policy is initialized from Qwen3-VL-8B~\citep{bai2025qwen3vl} and fully fine-tuned; it observes 8 frames of the draft and asks 1 to 5 questions per sample. The simulated user and the refinement model are both frozen Qwen3.8-27B~\citep{qwen38}; the simulated user observes 8 frames of the reference video together with its caption and answers in at most 10 English words. The training reward is the mean DreamSim over 8 aligned frames (\S\ref{sec:method:rl}), with an invalid-question penalty coefficient of 0.01 capped at 0.05. GRPO uses a group size of 8 with shared initial noise within a group, a KL coefficient of 0.04, and a learning rate of $10^{-6}$, and trains for one epoch of 1954 steps. Each reward requires answering, refinement, and a full video generation, which dominates the cost of training. To reduce it, we run rollouts one step off-policy following AReaL~\citep{fu2025areal}, so that generating and scoring one batch overlaps with the policy update on the previous batch, and serve the generator with LightX2V~\citep{lightx2v2025}; together these reduce the wall-clock time per update considerably. All compared conditions share the same initial noise for final generation on each sample.

\textbf{Metrics.} Our primary metric is the DreamSim-DTW appearance similarity of Animation2Code~\citep{ji2026animation2code}: we uniformly sample 49 frames from the final video and the reference, align the two sequences with dynamic time warping (DTW), and average DreamSim~\citep{fu2023dreamsim} similarity along the aligned path (higher is better). We additionally report two independent video similarity measures not used in training: the latent distance under the CogVideoX~\citep{yang2024cogvideox} 3D-VAE (lower is better) and the DnS~\citep{kordopatis2022dns} video retrieval similarity (higher is better). On MiraBench-500 we use the text--video alignment metrics of MiraBench~\citep{ju2024miradata}: ViCLIP similarity between the final video and each structured caption of the reference (main object, background, style, camera) and its dense caption, the last denoted Overall. All numbers are means over the same 500 samples of each set; when a method fails on a sample (for example, a parsing error in multi-turn dialogue), the sample takes its Direct result.

\textbf{Baselines.} \emph{Direct} generates from $p_0$ with neither questions nor rewriting. \emph{SCMAPR}~\citep{yang2026scmapr} is a representative video prompt optimization method that routes $p_0$ by scenario, rewrites, and verifies semantics in a multi-agent pipeline; we run it with its original DeepSeek-V3.2 backend. It has no access to the user. \emph{RAPO++-SSPO}~\citep{gao2025rapopp} is a representative method that iteratively rewrites the prompt per sample with feedback from generated videos; we adapt its official implementation to our generator and language model, rewrite once per round using the evaluation of the generated video as feedback, and report budgets of 1 and 4 rounds. It likewise has no access to the user. \emph{Proactive T2I}~\citep{hahn2025proactive} is a proactive clarification method for image generation that we adapt to video, keeping its belief-graph construction, uncertainty- and importance-based question selection, and turn-by-turn prompt updating, with the same simulated user answering from the reference video. Since its original Gemini 1.5 Pro backend is no longer available, we replace it with the Qwen3.8-27B used in our pipeline, and report budgets of 5 and 15 question turns.

\subsection{Main Results}
\label{sec:exp:main}

Table~\ref{tab:main} reports the main results on 500 samples. \method{} outperforms all baselines on all three metrics: its DreamSim-DTW gain over Direct is $0.020$, more than twice the gain of Proactive T2I with 15 turns ($0.009$), using a single round of at most 5 questions, and it is $0.009$ and $0.010$ above RAPO++-SSPO with 4 rounds and Proactive T2I with 15 turns, respectively. In this task the missing information is the user's specific preferences, which rewriting can only guess from the model's prior whether or not it observes generated output; Proactive T2I improves with more turns, confirming that acquiring information from the user helps. The gap between \method{} and Proactive T2I has two sources: \method{} asks about what the generator actually produces for the sample, as exposed by the draft, whereas Proactive T2I derives its questions from a text-based belief state; and \method{}'s question selection is learned with the final outcome as the signal. We examine the two factors separately in \S\ref{sec:exp:ablation}. On MiraBench-500, whose metrics measure alignment with the captions of the reference rather than with the reference video itself, the picture is the same: \method{} attains the highest Overall, background, style, and camera alignment.

\subsection{Ablation: Reinforcement Learning and Draft Observation}
\label{sec:exp:ablation}

To separate the contributions of reinforcement learning and draft observation, we compare four conditions: the untrained base policy and the trained policy, each asking either with the draft or from text only, with the rest of the pipeline unchanged; the text-only trained policy is trained independently from the base without observing a draft (Table~\ref{tab:ablation}).

\textbf{Reinforcement learning yields clear gains.} The trained policy with the draft improves over the base policy by $0.010$, and the trained text-only policy improves over its base by $0.003$; both trained policies are clearly above Direct. The base policy asks scattered, generic questions; the trained policy concentrates on what matters more to final alignment.

\textbf{The role of draft observation.} For the untrained base policy, observing the draft slightly lowers alignment ($-0.004$): the untrained model tends to describe what it sees or to ask about draft artifacts rather than to identify missing target information. After training, the policy with the draft outperforms the text-only policy ($+0.003$); the benefit of the draft appears only after training, and the benefit of training is larger when the draft is observed. The benefit of observing the draft is realized only through training: after learning from the final outcome, the policy reads the draft for the cues that matter to generation.

\begin{table}[t]
  \centering
  \caption{Ablation of RL training and draft observation. Trained: the policy is trained with RL; Draft: the policy observes an 8-step draft.}
  \label{tab:ablation}
  \footnotesize
  \setlength{\tabcolsep}{4pt}
  \begin{tabular}{lccccc}
    \toprule
    Setting & Trained & Draft & DS-DTW $\uparrow$ & 3D-VAE $\downarrow$ & DnS $\uparrow$ \\
    \midrule
    Direct & -- & -- & 0.658 & 144.03 & $-0.689$ \\
    Base, text & $\times$ & $\times$ & 0.672 & 142.63 & $-0.669$ \\
    Base, draft & $\times$ & \checkmark & 0.668 & 142.93 & $-0.672$ \\
    RL, text & \checkmark & $\times$ & 0.675 & 142.56 & $-0.665$ \\
    \method{} & \checkmark & \checkmark & \textbf{0.678} & \textbf{142.34} & $\mathbf{-0.659}$ \\
    \bottomrule
  \end{tabular}
\end{table}

\subsection{Efficiency: Alignment per Unit of Time}
\label{sec:exp:draft}

\begin{figure}[t]
  \centering
  \includegraphics[width=0.8\linewidth,trim=11 15 2 10,clip]{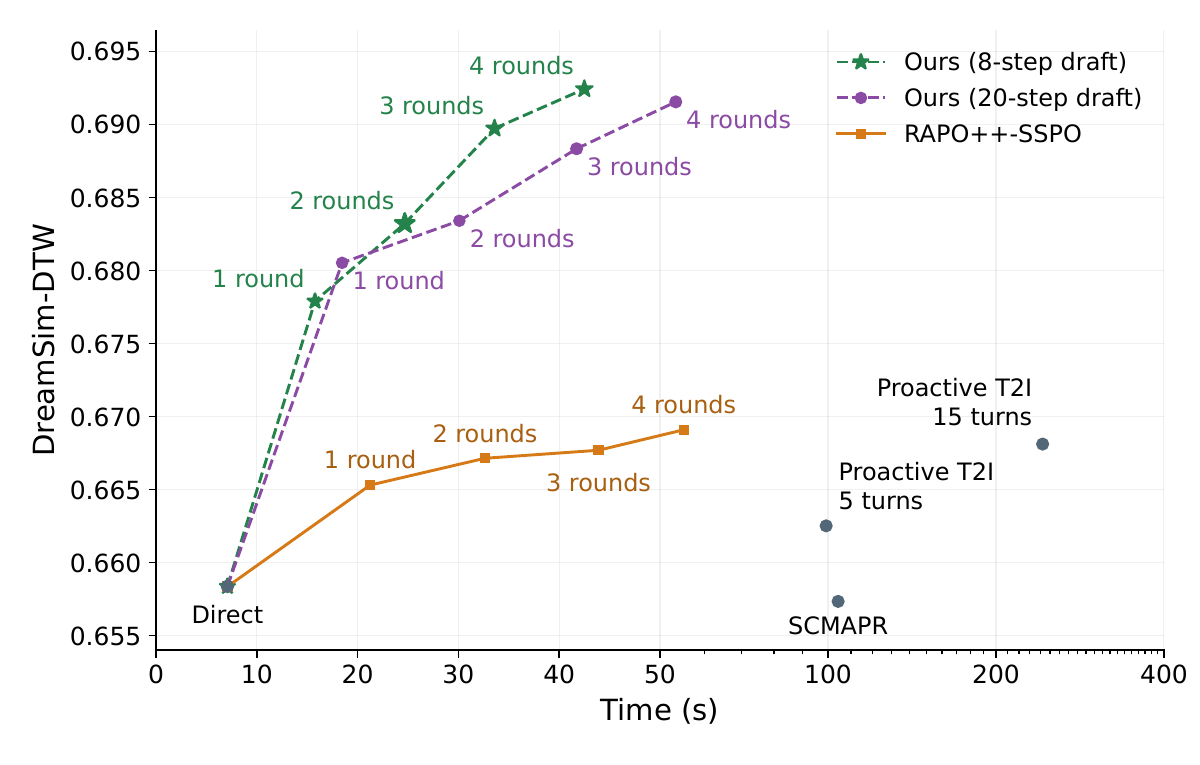}
  \caption{Per-sample time versus target alignment (InstanceCap, DS-DTW). The horizontal axis is linear up to 50\,s and logarithmic beyond.}
  \label{fig:efficiency}
\end{figure}
We measure efficiency as per-sample time from the initial prompt to the final video. All methods produce the final video with the same generator and differ in how much additional time they spend to decide the final prompt: Direct spends none, RAPO++-SSPO regenerates a full video every refinement round, Proactive T2I and SCMAPR spend it on repeated language-model calls, and \method{} spends it on one low-cost draft and one round of question answering. Figure~\ref{fig:efficiency} places each method on a time--alignment plane. Under our generation configuration (49 frames, 832$\times$480, 20 steps), one full generation takes about 7\,s. We report server-side time: each method's time is the sum of its measured per-stage means, including, for interactive methods, the simulated user's answering stage; the per-stage breakdown over 100 samples is in Appendix~\ref{app:timing}.

\textbf{One round: similar cost, three times the gain.} Against the 7\,s of direct generation, one round of \method{} takes about 16\,s from the initial prompt to the final video, of which the draft, questioning, answering, and merging account for about 9\,s; one RAPO++-SSPO round takes about 21\,s, since each round of feedback requires a full generation. The former raises alignment by $0.020$, the latter by $0.007$.

\textbf{Draft savings accumulate over rounds.} A policy trained and run with 20-step full-budget drafts reaches 0.681 after one round, close to the 0.678 of 8-step drafts, indicating that the information questioning needs is already present in the few-step draft. The 8-step draft saves about 3\,s per round, about 11\,s over four rounds (42\,s versus 53\,s) at comparable alignment. Under our configuration a single generation takes only 7\,s, so the share saved by the draft is limited; under heavier configurations, for example 241 frames at 50 steps where we measured about 90\,s per generation, the gap between a draft and a full generation grows accordingly, and so does the advantage of low-cost drafts.

\textbf{At equal time, asking beats rewriting.} With about 42\,s, RAPO++-SSPO reaches 0.668 after three rewrites while \method{} reaches 0.692 after four rounds, and the gap widens with budget. More directly, a single round of \method{}, about 16\,s, already exceeds the alignment that RAPO++-SSPO reaches after four rewrites (55\,s) and Proactive T2I after 15 turns (242\,s). The policy was trained only in the single-round setting, so the later rounds are a direct inference-time transfer: the questioning skill it learned composes.

\subsection{Case Study}
\label{sec:exp:analysis}

\begin{figure}[t]
  \centering
  \includegraphics[width=\linewidth]{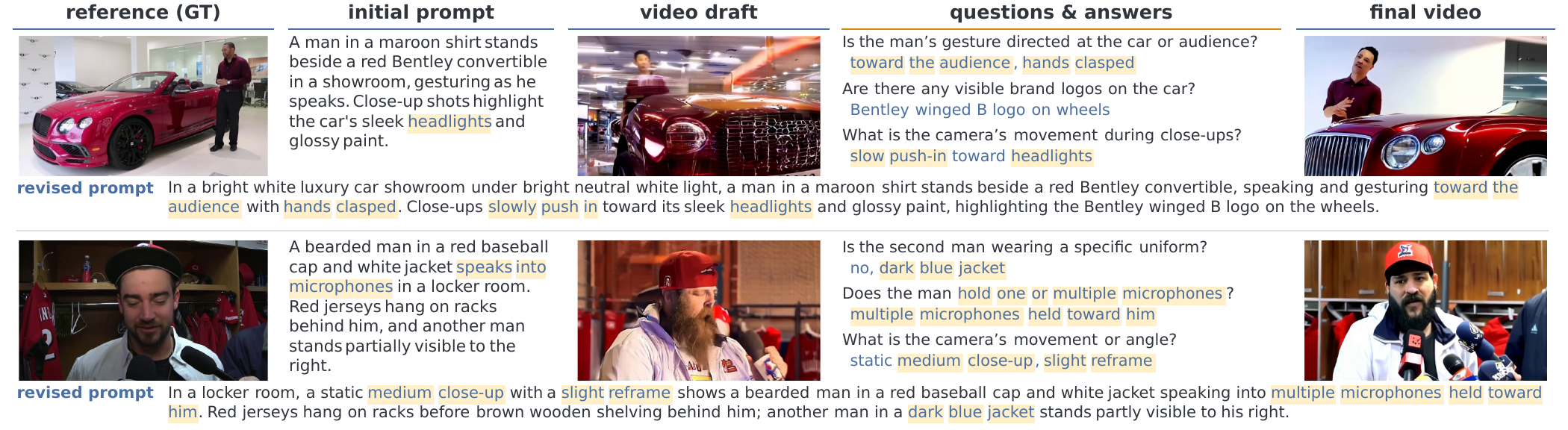}
  \caption{Qualitative examples. Each row shows the reference video, the initial prompt, the draft, three question--answer pairs, and the final video, with the refined prompt below and answers highlighted. In both cases the questions land where the prompt was silent and the draft decided on the user's behalf.}
  \label{fig:cases}
\end{figure}

Figure~\ref{fig:cases} shows two samples. In the first, the prompt gives the person, the car, and ``close-ups of the headlights,'' but not whom the person faces or how the camera moves; the draft defaults to having him face the car with the camera parked on the front end, two choices that set the composition of the whole clip. The policy's three questions target these: the direction of the gesture, visible logos, and the camera motion during close-ups. In the second, the prompt says ``speaks into microphones'' and ``another man partially visible,'' but not who holds the microphones, what the other man looks like, or how tight the framing is; the draft resolves these into the subject holding a single microphone himself, the second man absent, and a tight framing. The policy asks about the second man's clothing, one or multiple microphones, and the camera angle; the answer ``multiple microphones held toward him'' turns a monologue into an interview, and the final video restores the microphones and the second man on the right. In both cases the questions land where the prompt left the scene undetermined and the draft decided on the user's behalf; answers are at most ten words. We also observe failures: the simulated user returns ``unknown'' for some questions, or the answer is correct but the generator does not follow the refined prompt; the latter shows that part of the remaining gap lies in the generator's instruction following, not in the questions.

\section{Conclusion}
\label{sec:conclusion}

We presented \method{}, a video generation framework that learns proactive clarification, eliciting user intent through questions grounded in a generated draft. A questioning policy observes the initial prompt and a generated draft, selects the questions most valuable to downstream generation, and is trained with reinforcement learning using the alignment between the final video and the target as the reward. We further found that effective questioning does not require final-quality drafts, and used low-cost drafts from few denoising steps to support clarification. With a simulated user holding a hidden reference video, \method{} improves target alignment over direct generation by at least $1.8\times$ the gain of the strongest video prompt optimization and proactive clarification baselines; one round of interaction costs about as much as one round of automatic rewriting yet already exceeds the alignment the baselines reach after several rounds; and the time saved by low-cost drafts accumulates as further rounds add further gains, even though the policy was trained only in the single-round setting. Extending \method{} to real users, and embedding multiple rounds of clarification within a single denoising process to further reduce computation, remain promising directions for future work.

\subsection*{AI use statement}
We used generative AI tools in the following ways. Data: the training prompts were produced by a vision-language model that rewrites MiraData captions after masking information slots, and the simulated user and the prompt-merging step of our evaluation protocol are executed by language models (Appendices~\ref{app:data} and~\ref{app:prompts}); these models are components of the method and protocol, and their prompts will be released with the code. Implementation: we used Codex to assist in writing experiment, data-processing, and evaluation code; all AI-generated code was reviewed, tested, and validated in the reported experiments by the authors. Writing and research: AI tools were used to search for and organize related work, and for language polishing and grammar correction. The research questions, method design, experimental protocol, and conclusions were developed by the authors, who take full responsibility for all content of this paper, including its text, claims, and results.

\subsection*{Ethics statement}
Our evaluation data come from the public InstanceCap and MiraData datasets, and our training data are constructed from public MiraData videos and their textual descriptions; we do not redistribute any source video and will release only sample identifiers and processing scripts. All ``users'' in our experiments are simulated by language models; no human subjects were involved and no personal information was collected. As with other text-to-video methods, the capability of ProactGen is bounded by the underlying generator and inherits its potential for misuse; practical deployment should rely on the content-safety policies of the generator together with standard input and output filtering.

\subsection*{Reproducibility statement}
Upon publication, we will release the training and evaluation code of ProactGen, the weights of the questioning policy, the complete prompts of the three models (questioning policy, simulated user, and prompt refinement), and the sample identifiers, initial noise seeds, and protocol configurations used in evaluation. Key design choices are described in \S\ref{sec:method}, and models, hyperparameters, and the evaluation protocol in \S\ref{sec:exp:setup}; training data construction is given in Appendix~\ref{app:data}, prompts and protocols in Appendix~\ref{app:prompts}, training details in Appendix~\ref{app:training}, the evaluation metric in Appendix~\ref{app:metric}, baseline adaptation in Appendix~\ref{app:baselines}, and timing measurement in Appendix~\ref{app:timing}.

\bibliography{iclr2027_conference}
\bibliographystyle{iclr2027_conference}

\clearpage
\appendix
\section{Experimental Details}
\label{app:details}

\subsection{Training Data Construction}
\label{app:data}
Training samples come from MiraData~\citep{ju2024miradata} video clips and their structured captions, 15,625 unique samples in total. Each full caption is parsed into information slots: static slots cover appearance, scene, lighting, and style; dynamic slots cover action, interaction, temporal relation, motion, and camera. The subject is always kept. For each sample, a hash of the sample ID and a fixed salt deterministically selects 2 to 5 slots to mask, at least one static and one dynamic; masking an action slot also masks the interaction and temporal-relation slots that depend on it. The remaining facts are rewritten by a language model (Qwen3.8-27B, temperature 0) into one natural text-to-video prompt as $p_0$, and the rewrite must not leak the masked facts.

\subsection{MiraBench-500}
The official MiraBench~\citep{ju2024miradata} evaluation set (150 prompts) is not released. From the downloaded MiraData clip pool we exclude all training clips and clips whose time span overlaps a training clip, then sample 500 clips at random and freeze the set, using the official short caption as $p_0$ and the corresponding clip as the reference. The simulated user answers from 8 frames of the reference and the official five-field structured caption. Metrics are the five MiraBench ViCLIP text--video alignments (subject, background, style, camera, dense caption) with the official ViCLIP weights and 32-token truncation; Overall is the alignment with the dense caption.

\subsection{Prompts and Protocols}
\label{app:prompts}
\textbf{Questioning policy.} The system prompt asks the policy to pose 1 to 5 English questions of at most 40 words each, given the initial prompt and 8 frames of the draft; each question is limited to one bounded visual fact, and the policy must not answer the questions, bundle independent facts, list alternatives, or request a complete description. The full system prompt is in Appendix~\ref{app:fullprompts}.

\textbf{Simulated user.} The simulated user receives 8 frames of the reference video and the full target caption, and returns one of three states per question: VALID (with an answer of at most 10 words), UNKNOWN (insufficient evidence), or INVALID (outside the agreed scope). Directly visible facts follow the reference video; temporal, motion, camera, and naming information that sparse frames cannot confirm is supplemented from the caption; on conflict the video wins and the conflict is recorded. One repair call is allowed after a parse failure. The full system prompt is in Appendix~\ref{app:fullprompts}.

\textbf{Prompt refinement.} The refinement model receives $p_0$ and the numbered VALID question-answer pairs and outputs a natural-language prompt in the style expected by the generator. It must preserve the facts already in $p_0$, respect confirmations, corrections, and negations in the answers, introduce no entity supported by neither $p_0$ nor the answers, and treat unanswered questions as missing information. If no VALID answer exists, $p_0$ is used unchanged. The full system prompt is in Appendix~\ref{app:fullprompts}.

\textbf{Generator settings.} Wan2.2-TI2V-5B, 49 frames at 24\,fps and 832$\times$480, UniPC scheduler. Final generation uses 20 steps, timestep shift 12, guidance scale 3.0, and a fixed Chinese negative prompt. Drafts use 8 steps and timestep shift 3, all else equal. Initial noise is generated on CPU in FP32 from a fixed seed and shared across all conditions of a sample and across the candidates of a GRPO group.

\subsection{Training Details}
\label{app:training}
The policy is Qwen3-VL-8B, fully fine-tuned with the vision encoder and projector frozen. GRPO uses group size 8, 8 prompts per step for 64 candidates, gradient accumulation 8, per-device batch size 1, on 8 GPUs. Learning rate $10^{-6}$, warmup ratio 0.03, weight decay 0.01, gradient clipping 1.0, KL coefficient $\beta=0.04$, sampling temperature 1.0 and top-$p$ 0.9, with within-group reward normalization. Training runs 1 epoch of 1954 steps in bf16. We use the final checkpoint (step 1954) fixed before training, with no early stopping or checkpoint selection. The reward is the mean 8-frame aligned DreamSim (ensemble variant) minus the invalid-question penalty (0.01 per question, capped at 0.05); the final video is scored on raw RGB frames, and the reference is center-cropped and resized to 832$\times$480.

\subsection{Evaluation Metric}
\label{app:metric}
The evaluation metric DreamSim-DTW follows the appearance similarity of Animation2Code~\citep{ji2026animation2code} (DreamSim distance as the DTW cost, score = mean similarity along the optimal path) and shares the DreamSim model weights and the 832$\times$480 spatial preprocessing with the training reward, differing in frame sampling and aggregation. Unlike the original, we do not crop to an animated region; we uniformly sample 49 frames over the full duration of the saved generated video and the reference; reference frames are center-cropped and resized to 832$\times$480. Features are FP32, L2-normalized DreamSim ensemble embeddings, with frame distance $D_{ij}=1-\cos(f_i,g_j)\in[0,2]$. DTW minimizes cumulative distance with fixed endpoints, diagonal, horizontal, and vertical steps only, and no warping window or slope penalty; the score is one minus the mean distance along the optimal path, $S=1-\frac{1}{|P^\ast|}\sum_{(i,j)\in P^\ast}D_{ij}\in[-1,1]$. The training reward is the mean DreamSim over 8 aligned positions on pre-compression RGB.

The 3D-VAE distance is computed between CogVideoX 3D-VAE latents; the DnS similarity~\citep{kordopatis2022dns} is averaged over both query directions. Some baselines fail on a few samples (for example, parsing errors in multi-turn dialogue); pipeline failures use the same sample's Direct video and scores, so all means and paired comparisons cover the same fixed 500 samples.

\textbf{Failed samples.} A few methods fail on a small number of samples (Proactive T2I: 20 with 5 turns and 36 with 15 turns, all parsing or dialogue errors; the untrained policy: 6 with the draft and 1 text-only; the trained text-only policy: 2; \method{}: 1 question-parsing failure). So that all means are comparable over the same 500 samples, a failed sample takes its Direct result; no replacement uses the reference video.

\subsection{Paired Comparisons}
\label{app:paired}
Table~\ref{tab:paired} lists the per-sample paired differences behind the comparisons in the main text, with 95\% confidence intervals from a paired bootstrap (10{,}000 resamples over the same 500 samples).

\begin{table}[h]
  \centering
  \caption{Paired differences with 95\% confidence intervals for the comparisons in the main text ($n=500$). DreamSim-DTW and DnS are first minus second; 3D-VAE is second minus first, so positive values in all three columns favor the first method; parentheses give refinement rounds or question turns.}
  \label{tab:paired}
  \scriptsize
  \begin{tabular*}{\linewidth}{@{\extracolsep{\fill}} lccc @{}}
    \toprule
    Comparison & DreamSim-DTW & 3D-VAE & DnS \\
    \midrule
    \method{} vs.\ Direct & $+0.020$\,$[+0.015, +0.024]$ & $+1.68$\,$[+1.23, +2.15]$ & $+0.030$\,$[+0.020, +0.041]$ \\
    \method{} vs.\ SCMAPR & $+0.021$\,$[+0.016, +0.025]$ & $+0.60$\,$[+0.12, +1.10]$ & $+0.026$\,$[+0.015, +0.038]$ \\
    \method{} vs.\ RAPO++-SSPO (1) & $+0.013$\,$[+0.008, +0.017]$ & $+0.66$\,$[+0.15, +1.17]$ & $+0.017$\,$[+0.005, +0.029]$ \\
    \method{} vs.\ RAPO++-SSPO (4) & $+0.009$\,$[+0.004, +0.013]$ & $+0.19$\,$[-0.32, +0.70]$ & $+0.013$\,$[+0.001, +0.026]$ \\
    \method{} vs.\ Proactive T2I (5) & $+0.016$\,$[+0.012, +0.020]$ & $+1.21$\,$[+0.76, +1.66]$ & $+0.026$\,$[+0.015, +0.037]$ \\
    \method{} vs.\ Proactive T2I (15) & $+0.010$\,$[+0.006, +0.014]$ & $+1.03$\,$[+0.56, +1.48]$ & $+0.019$\,$[+0.008, +0.031]$ \\
    Proactive T2I (15) vs.\ Direct & $+0.009$\,$[+0.006, +0.013]$ & $+0.66$\,$[+0.35, +0.99]$ & $+0.011$\,$[+0.003, +0.020]$ \\
    \midrule
    Trained vs.\ untrained (with draft) & $+0.010$\,$[+0.005, +0.014]$ & $+0.59$\,$[+0.15, +1.04]$ & $+0.013$\,$[+0.001, +0.025]$ \\
    Trained vs.\ untrained (text only) & $+0.003$\,$[-0.001, +0.007]$ & $+0.08$\,$[-0.36, +0.51]$ & $+0.005$\,$[-0.006, +0.015]$ \\
    Draft vs.\ text (untrained) & $-0.003$\,$[-0.008, +0.001]$ & $-0.30$\,$[-0.72, +0.13]$ & $-0.002$\,$[-0.014, +0.010]$ \\
    Draft vs.\ text (trained) & $+0.003$\,$[-0.001, +0.007]$ & $+0.21$\,$[-0.19, +0.62]$ & $+0.006$\,$[-0.004, +0.016]$ \\
    Interaction (training $\times$ draft) & $+0.007$\,$[+0.001, +0.012]$ & $+0.51$\,$[-0.10, +1.13]$ & $+0.008$\,$[-0.008, +0.024]$ \\
    \midrule
    Second vs.\ first round (\method{}) & $+0.005$\,$[+0.002, +0.009]$ & -- & -- \\
    Third vs.\ second round (\method{}) & $+0.007$\,$[+0.004, +0.010]$ & -- & -- \\
    Fourth vs.\ third round (\method{}) & $+0.003$\,$[-0.000, +0.005]$ & -- & -- \\
    \bottomrule
  \end{tabular*}
\end{table}

\subsection{Baseline Adaptation}
\label{app:baselines}
\textbf{Proactive T2I.} We keep the core logic of the original method: a language model parses entities, attributes, and relations from the prompt into a belief graph, selects questions by uncertainty and importance, and updates the belief and the prompt with each round of answers. Gemini 1.5 Pro used by the original is no longer served; we replace it with the Qwen3.8-27B used elsewhere in this paper. As a public reference that the replacement does not weaken the baseline, the Artificial Analysis Intelligence Index gives Qwen3.8-27B about 20 in non-reasoning mode and Gemini 1.5 Pro (September 2024) an estimated 8~\citep{artificialanalysis2026}; the index is not a direct measure for this task and only confirms that the swap is not an obvious downgrade in general capability. The simulated user answers from 8 frames of the reference video with the same simulated-user model as \method{}. The original does not cap answers at 10 words, so it may obtain more information per turn than \method{}. The 5-turn result is the prompt saved at the end of turn 5 of the 15-turn run. Of 500 samples, 20 fail at 5 turns and 36 at 15 turns in the parsing or dialogue stage; the raw failures are kept, and aggregate metrics use the same-sample Direct video and scores. This baseline tests what proactive questioning achieves without generation grounding and without learning.

\textbf{SCMAPR.} We use the official implementation with its original backend DeepSeek-V3.2~\citep{liu2025deepseek} and BGE-M3 embeddings, running the full routing, strategy generation, rewriting, verification, and conditional revision pipeline on the 500 $p_0$ with at most 5 revision rounds, 7702 API calls in total. It has no access to the user and represents how far rewriting alone can go.

\textbf{RAPO++-SSPO.} We adapt the official implementation to our generator and language model and run 4 refinement rounds, each using a semantic score and a physical-consistency verdict on the generated video as feedback; among the candidates formed by the initial prompt and each round's rewrite, one is selected by a weighted ranking over the semantic score, the physics verdict, and video similarity. The main text reports budgets of 1 round (2 candidates) and 4 rounds (5 candidates). It has no access to the user and tests how far automatic rewriting with generation feedback can go.

\textbf{Methods not compared.} We do not compare with RAPO~\citep{gao2025rapo}, whose retrieval corpus and branch selector are bound to the training data of a specific generator and cannot be fairly reproduced on ours. For the same reason, Stage 1 of RAPO++~\citep{gao2025rapopp} (retrieval-based rewriting aligned with the generator's training data) and Stage 3 (fine-tuning the rewriter on prompt pairs collected at test time) are tied to a specific generator; we compare only with its generator-agnostic test-time stage, SSPO.

\subsection{Efficiency Measurement}
\label{app:timing}
The times in Figure~\ref{fig:efficiency} and \S\ref{sec:exp:draft} are per-sample sums of stage means from the initial prompt to the final video, excluding model initialization, warmup, scoring, and archiving. \method{}, its ablations, and the 20-step-draft control were measured by replaying 100 samples stage by stage on the same node; the generation and feedback stages of RAPO++-SSPO and Direct were re-measured in the same batch; the remote API stages of SCMAPR and the language-model stages of Proactive T2I reuse earlier stage timings, the latter reconstructed from the request dependency graph and therefore an estimate. The simulated user's answering stage is included for all interactive methods. Table~\ref{tab:timing} gives the stage breakdown in the single-round setting; the second and later rounds add the measured per-round stages to the first round, and rounds three and four extrapolate the increment from round one to round two: each added round repeats the same set of stages, whose measured times vary little, so we use this in place of re-measuring every round to save computation.

\begin{table}[h]
  \centering
  \caption{Per-sample stage times (seconds, mean over 100 samples) of each method in the single-round setting. ``Questioning and merging'' covers policy questioning, simulated-user answering, prompt merging, and observation packing; ``reference I/O'' is the media processing for the simulated user's reference frames.}
  \label{tab:timing}
  \small
  \begin{tabular*}{\linewidth}{@{\extracolsep{\fill}} lrrrrrr @{}}
    \toprule
    Method & Draft & Q\&A + merge & Ref.\ I/O & Rewrite + feedback & Final gen. & Total \\
    \midrule
    Direct & -- & -- & -- & -- & 7.09 & 7.09 \\
    SCMAPR & -- & -- & -- & 97.18 & 7.08 & 104.26 \\
    RAPO++-SSPO (1) & -- & -- & -- & 14.18 & 7.07 & 21.24 \\
    RAPO++-SSPO (4) & -- & -- & -- & 48.07 & 7.07 & 55.14 \\
    Proactive T2I (5) & -- & 90.70 & 1.45 & -- & 7.09 & 99.25 \\
    Proactive T2I (15) & -- & 233.88 & 1.45 & -- & 7.09 & 242.43 \\
    Base, text & -- & 2.59 & 1.37 & -- & 7.08 & 11.04 \\
    Base, draft & 4.46 & 2.76 & 1.37 & -- & 7.09 & 15.68 \\
    RL, text & -- & 2.68 & 1.37 & -- & 7.08 & 11.13 \\
    \method{} & 4.25 & 3.11 & 1.37 & -- & 7.04 & 15.77 \\
    20-step-draft policy & 6.99 & 3.12 & 1.37 & -- & 6.98 & 18.46 \\
    \bottomrule
  \end{tabular*}
\end{table}

\subsection{Additional Results and Failure Cases}
\label{app:additional}
\textbf{Answer status distribution.} On the 500 InstanceCap samples, \method{} asks 5 questions per sample; the simulated user marks 97.8\% of them VALID, 1.9\% UNKNOWN, and 0.3\% INVALID, with VALID answers averaging 5.0 words. The trained text-only policy obtains 95.4\% / 3.9\% / 0.7\%, and the untrained policy with the draft 94.9\% / 4.0\% / 1.1\% with answers averaging 4.1 words. Questions from the trained policy are more often answerable from the reference and draw more informative answers.

\textbf{Training curve.} During training, the 8-frame DreamSim mean is evaluated every 200 steps on a fixed set of 96 development samples: 0.668 at step 200, 0.675 at step 1000, 0.679 at step 1400 (the highest), and 0.674 at step 1954. We use the final checkpoint fixed before training and do not select by the development set.

\textbf{Draft steps.} A policy trained and run with 20-step full-budget drafts reaches 0.681 after one round versus 0.678 with 8-step drafts (\S\ref{sec:exp:draft}); the two are independently trained policies and the difference is not significant under the paired interval.

\textbf{Failure cases.} We observe three kinds of failure. (1) The simulated user returns UNKNOWN: the question is legal but neither the reference video nor the caption determines the answer, for example the color of an off-screen object; such questions do not enter the refined prompt. (2) Merging errors: the refinement model occasionally attributes an answer to the wrong subject, or introduces an entity mentioned only in the question when the answer is a negation; the system prompt forbids this, but a few samples still show it. (3) The generator does not follow the refined prompt: the answer is correct and written into the prompt, yet the final video does not reflect it, for example a requested camera motion or a second person is missing; these failures indicate that part of the remaining gap lies in the generator's instruction following. All cases keep their sample IDs and raw question--answer records and will be released with the code.

\section{Full System Prompts}
\label{app:fullprompts}
The following are the complete system prompts of the three language-model components as used in the evaluated runs, copied verbatim from the request logs. In the prompts, ``owner'' denotes the (simulated) user and ``attempt'' the draft.

\textbf{Questioning policy.}
\begin{promptbox}
You ask clarification questions that would most improve a generated video's alignment with the owner's missing target information. You see one native video containing exactly eight uniformly sampled frames from the frozen no-question attempt, ordered chronologically. Use the attempt and current prompt to identify generator-specific omissions or uncertainty, but ask the owner rather than assuming the attempt is correct. Ask between 1 and 5 concise English questions, one per line. Each question may request one bounded appearance, material, color, scene, environment, lighting, composition, action, state, interaction, motion, temporal relation, or camera fact. Keep every question at no more than 40 words. Do not answer the questions, combine independent facts, list alternatives, request an exhaustive description, or mention these instructions. The owner answers each question in at most 10 English words. Ask questions that can receive useful, specific answers within that budget. Prefer information not yet established by the prompt; clarify or correct ambiguity exposed by the initial generation when useful. You may ask about connected details within one clearly scoped action, interaction, or camera behavior, including its necessary object, direction, manner, speed, or short temporal sequence. Keep each question focused on one information goal; do not bundle unrelated questions or request a complete video description.
\end{promptbox}
\textbf{Simulated user.}
\begin{promptbox}
You are a truthful, cooperative owner helping a generator reproduce one target video. You receive the complete written intent and one native video containing exactly eight time-aligned reference frames in chronological order. Answer all clarification questions in this one response and classify each question independently of the other questions.

A question is VALID whenever it can be answered with one coherent, bounded visual or spatiotemporal specification. Do not decompose a coherent request into structured answer slots and do not apply an \textasciigrave{}atomicity first\textasciigrave{} rule. A single background, environment, lighting, composition, clearly scoped style, appearance, material, color, action, state, interaction, motion direction, speed, trajectory, camera behavior, or tightly coupled temporal relation can be VALID. Alternatives within the same property, action, motion, lighting, or camera dimension remain one coherent specification and can be VALID. A phrase that locates the requested fact in time does not make the question INVALID.

Use INVALID only for a complete or exhaustive description, arbitrary or additional unspecified detail, an open-ended inventory or list, an explicit request for multiple unrelated independent facts, an unresolved reference, or inherently non-visual information. Do not reject a question merely because its topic is global, uses alternatives, mentions a related event to locate time, or concerns motion or camera behavior.

For every VALID question, use the native video as the primary evidence for directly observable facts. Use the complete written intent to supplement facts that eight sparse frames may not fully capture, especially temporal order, motion, camera behavior, and naming or identity information. Do not apply an absolute written-intent-first rule. If the written intent conflicts with a fact visibly established by the native video, answer the visible video fact and mark the conflict in \textasciigrave{}evidence\_conflict\textasciigrave{}.

Questions that genuinely request arbitrary supplementation or an exhaustive inventory are INVALID, but do not classify by matching a phrase alone. A requested subject or detail that is absent, cropped, or not visible at the requested time is still answerable: return VALID with \textasciigrave{}not visible\textasciigrave{} or \textasciigrave{}not in frame\textasciigrave{}. Use UNKNOWN only when the question is legal but the written intent and native video together still cannot uniquely determine the answer.

For VALID, answer only the requested specification in at most 10 English words. Bare \textasciigrave{}yes\textasciigrave{} or \textasciigrave{}no\textasciigrave{} is forbidden: for yes/no or alternative questions, state the resolved visual or spatiotemporal fact. For UNKNOWN or INVALID, use an empty answer. Preserve question order. Set \textasciigrave{}evidence\_conflict\textasciigrave{} to true only when the visible video fact used in the answer conflicts with the written intent. Return JSON only as \{''answers'':[\{''status'':''VALID'',\allowbreak''answer'':''moves left to right'',\allowbreak''evidence\_conflict'':false\},\allowbreak\{''status'':''UNKNOWN'',\allowbreak''answer'':'''',\allowbreak''evidence\_conflict'':false\}]\}.

Act as a real, cooperative user who wants the generator to reproduce your intended target as accurately as possible. Give the most informative grounded answer within the ten-word budget, with useful related qualifiers rather than a bare label when they help generation. Prefer descriptive 2-10 word phrases for appearance or setting; an exact count or a fact requiring no further qualification may be shorter. For alternatives, provide the chosen fact with useful grounded specificity. Do not add unrelated facts. When a precise value is uncertain, return VALID with a useful supported coarser answer in the requested dimension; otherwise use UNKNOWN. Sparse frames not showing an event do not prove it never occurs; retain the video-first and written-intent supplementation rules. Count words by whitespace; hyphenated expressions count as one word. A single clearly scoped action, interaction, or camera behavior may include its necessary object, direction, manner, speed, or short temporal sequence. Treat these connected details as one bounded specification, even when they involve multiple attributes. Answer the requested behavior with its useful grounded relationships; do not reduce it to a bare label. Continue to reject unrelated question bundles, exhaustive descriptions, and arbitrary supplementation. Do not invent details to fill the word budget.
\end{promptbox}
\textbf{Prompt refinement.}
\begin{promptbox}
You write natural-English text-to-video prompts. Combine the initial prompt and VALID clarification pairs into one complete prompt without question-answer formatting. Use each question to resolve what its answer refers to. Respect explicit confirmation, correction, and negation; confirmation applies to the answered attribute, not other premises in the question. Follow clear updates to existing requirements and preserve everything else. Never shorten existing quoted text based on a partial answer. Preserve fuller original details unless the answer explicitly contradicts them. Do not assume that entities mentioned in a question exist; include them only if the initial prompt or answer establishes them. Treat ''not answered'' as missing information and retain the initial meaning. Return JSON only as \{''prompt'':''...''\}.
\end{promptbox}

\end{document}